\documentclass[sigconf]{acmart}

\newtheorem{problem}{Problem}
\usepackage{multirow}
\usepackage{caption}
\usepackage{subcaption}
\usepackage{float}
\usepackage{ulem}
\AtBeginDocument{%
  }

\setcopyright{acmlicensed}
\copyrightyear{2025}
\acmYear{2025}
\acmDOI{XXXXXXX.XXXXXXX}
\acmConference[ICAIL '25]{Proceedings of the 20th International Conference on Artificial Intelligence and Law}{June 16--20, 2025}{Chicago, IL}

\acmISBN{978-1-4503-XXXX-X/18/06}

\begin{document}

\title{Universal Legal Article Prediction via Tight Collaboration between Supervised Classification Model and LLM}

\author{Xiao Chi}
\email{cx3506@zju.edu.cn}
% \orcid{0009-0005-9902-1577}
\affiliation{%
  \institution{Zhejiang University}
  \city{Hangzhou}
  \state{Zhejiang}
  \country{China}
}

\author{Wenlin Zhong}
\email{m13527860108@163.com}
\orcid{}
\affiliation{%
  \institution{Zhejiang University}
  \city{Hangzhou}
  \state{Zhejiang}
  \country{China}
}

\author{Yiquan Wu}
\authornote{Corresponding author.}
\email{wuyiquan@zju.edu.cn}
\orcid{}
\affiliation{%
  \institution{Zhejiang University}
  \city{Hangzhou}
  \state{Zhejiang}
  \country{China}
}

\author{Wei Wang}
\email{wei031221@163.com}
\orcid{}
\affiliation{%
  \institution{Zhejiang University}
  \city{Hangzhou}
  \state{Zhejiang}
  \country{China}
}

\author{Kun Kuang}
\email{kunkuang@zju.edu.cn}
\affiliation{%
  \institution{Zhejiang University}
  \city{Hangzhou}
  \state{Zhejiang}
  \country{China}
}

\author{Fei Wu}
\email{wufei@zju.edu.cn}
\affiliation{%
  \institution{Zhejiang University}
  \city{Hangzhou}
  \state{Zhejiang}
  \country{China}
}

\author{Minghui Xiong\textsuperscript{*}}
\email{xiongminghui@zju.edu.cn}
\affiliation{%
  \institution{Zhejiang University}
  \city{Hangzhou}
  \state{Zhejiang}
  \country{China}
}

\begin{abstract}

Legal Article Prediction (LAP) is a critical task in legal text classification, leveraging natural language processing (NLP) techniques to automatically predict relevant legal articles based on the fact descriptions of cases. As a foundational step in legal decision-making, LAP plays a pivotal role in determining subsequent judgments, such as charges and penalties.
Despite its importance, existing methods face significant challenges in addressing the complexities of LAP. Supervised classification models (SCMs), such as CNN and BERT, struggle to fully capture intricate fact patterns due to their inherent limitations. Conversely, large language models (LLMs), while excelling in generative tasks, perform suboptimally in predictive scenarios due to the abstract and ID-based nature of legal articles. Furthermore, the diversity of legal systems across jurisdictions exacerbates the issue, as most approaches are tailored to specific countries and lack broader applicability.
To address these limitations, we propose Uni-LAP, a universal framework for legal article prediction that integrates the strengths of SCMs and LLMs through tight collaboration. Specifically, in Uni-LAP, the SCM is enhanced with a novel Top-K loss function to generate accurate candidate articles, while the LLM employs syllogism-inspired reasoning to refine the final predictions.
We evaluated Uni-LAP on datasets from multiple jurisdictions, and empirical results demonstrate that our approach consistently outperforms existing baselines, showcasing its effectiveness and generalizability.

\end{abstract}





\maketitle

\section{Introduction}
% 场景重要性
Recent advancements in neural networks and large-scale pre-trained language models have significantly expanded the capabilities of artificial intelligence (AI) across a wide range of domains. Within the legal field, Legal AI has emerged as a prominent area of interest, attracting growing attention from both academic researchers and practitioners. Its potential to improve the efficiency of legal practice and provide accessible legal assistance has become a focal point for innovation and development.
LAP is a task of predicting the relevant laws and regulations involved in a case according to the fact description text of the case, and it has broad application prospects in improving judicial efficiency. LAP can be regarded as a special text classification task, which connects case facts to subsequent legal judgments and is an essential step in both the civil law and common law systems \cite{DBLP:conf/nips/GuhaNHRCKCPWRZT23}. This step is crucial not only in legal judgment prediction (LJP), which applies machine learning techniques for automated legal reasoning, but also beneficial to another branch of research that focuses on legal cases reasoning based on knowledge \cite{bench2009argumentation, prakken2015formalization}. These methods reason about cases in a way similar to judges, as they focus on making decisions rather than predicting outcomes. As a result, they have strong logical reasoning abilities. However, these systems face challenges in acquiring knowledge such as rules, and data-driven methods for finding applicable legal articles could help address these issues.

\begin{figure*}[h]
\centering
\includegraphics[width=1\textwidth]{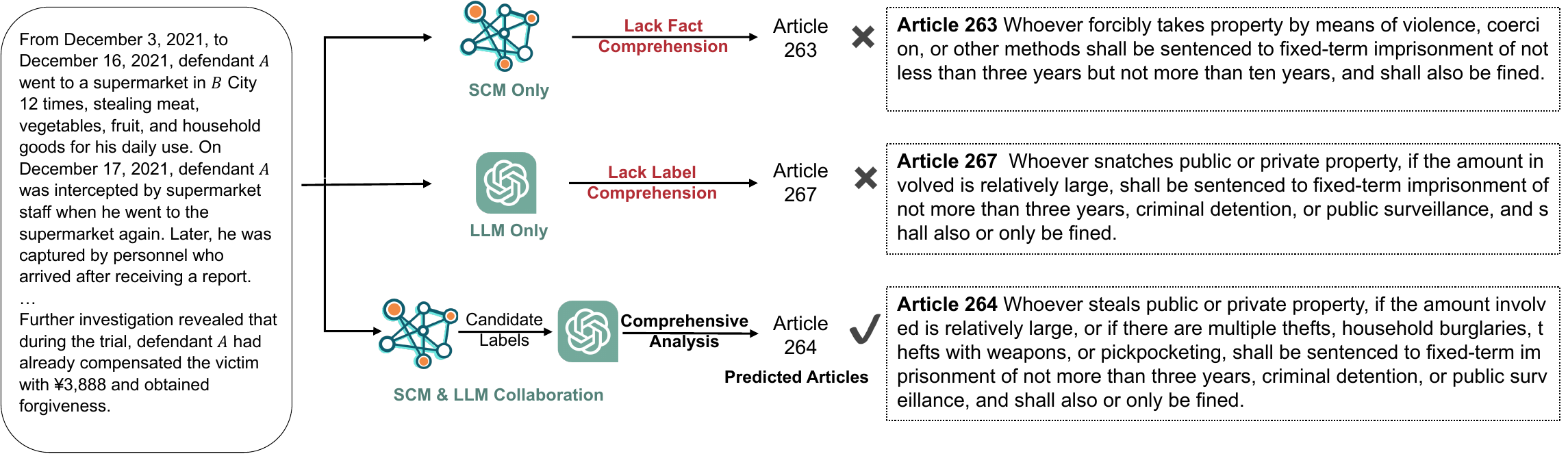}
\caption{An illustration of three different approaches to LAP on a Chinese legal case (translated). For a given fact description, SCM Only uses a supervised classification model to produce the output, while LLM Only relies on a LLM to generate the result directly. The Collaborate approach integrates both the SCM and the LLM to yield the final output.}
\label{fig1}
\end{figure*} 

% The goal of LJP is to predict judgment outcomes, including legal articles, charges, and terms of penalty, based on case fact descriptions. This paper focuses on Legal Article Prediction (LAP), which corresponds to rule-recall in the legal reasoning process\cite{ellsworth2005legal}. LAP connects case facts to subsequent legal judgments and is an essential step in both common law and civil law systems. \cite{DBLP:conf/nips/GuhaNHRCKCPWRZT23}. This step is crucial not only in LJP tasks, which apply machine learning techniques for automated legal reasoning, but also beneficial to another branch of research that focuses on reasoning about legal cases based on knowledge \cite{hage1996theory, bench2009argumentation, prakken2015formalization}. These methods reason about cases in a way similar to judges, as they focus on making decisions rather than predicting outcomes. As a result, they have strong logical reasoning abilities and achieve high accuracy. However, these systems face challenges in acquiring knowledge such as rules, and data-driven methods for finding matching legal articles could help address these issues.

% Describing the challenges in predicting law articles
% Challenges in legal article prediction 
The primary challenge of LAP lies in the inherent complexity and abstraction of legal articles. Compared to other legal text classification tasks, LAP demands a higher degree of precision and interpretative text analysis due to the unique characteristics of legal articles. These articles are not only logically structured to define legal responsibilities and obligations with precision but also require extensive interpretation to be applied across diverse contexts. Furthermore, they exhibit high heterogeneity, often encompassing complex syntax, specialized terminology, and culturally embedded nuances \cite{DBLP:conf/cikm/ShuZLDDZ24}.
Currently, research on LAP predominantly follows two main approaches. The first approach leverages SCMs \cite{DBLP:conf/sigir/LiuWZS0WK23, DBLP:journals/apin/ChenCW22}. These models are trained with annotated labels and demonstrate a strong ability to grasp abstract categories, enabling accurate identification of relevant legal articles \cite{DBLP:conf/emnlp/WuZL0LZS0K23}. However, their capacity to understand nuanced case details and capture intricate fact patterns is limited by their scale and architecture, making them less effective in handling ambiguous or complex cases.
The second approach involves the use of LLMs, which have recently gained significant attention in the Legal AI domain due to their remarkable text comprehension capabilities \cite{DBLP:conf/emnlp/Shui00C23, DBLP:journals/access/WangZHSZ24}. LLMs excel in understanding and generating coherent, human-like text in legal contexts, effectively handling specialized terminology and context-specific logical structures \cite{DBLP:journals/corr/abs-2307-06435}. However, evaluation studies reveal that LLMs perform poorly in predictive tasks \cite{DBLP:conf/emnlp/DengMZD24}, particularly in the LAP setting \cite{DBLP:conf/emnlp/WuZL0LZS0K23}. This underperformance stems from their reliance on prompt-based mechanisms, which struggle with the large number of abstract labels, such as unique IDs associated with legal articles.
Moreover, most existing approaches are typically designed for specific jurisdictions to tackle different legal systems across countries and regions, and therefore lack generalizability \cite{DBLP:conf/emnlp/WuZL0LZS0K23, DBLP:conf/iciss2/StricksonI20}.

To address the aforementioned challenges and enhance both the accuracy and efficiency of LAP, in this paper, we propose a universal framework for LAP, \textbf{Uni-LAP}, which leverages the complementary strengths of SCMs and LLMs as illustrated in Figure \ref{fig1}: SCMs provide candidate labels, while LLMs make the final decision. Unlike prior works that use SCMs and LLMs independently \citep{DBLP:conf/emnlp/WuZL0LZS0K23}, Uni-LAP emphasizes close collaboration between these two types of models by tackling two critical issues:

1) How can SCMs be trained to effectively integrate with LLMs?

2) How can LLMs be designed to account for the distinct characteristics of legal articles in prediction tasks?

For the first issue, Uni-LAP adopts a novel Top-K loss function for SCMs, which strategically prioritizes the accuracy of Top-K labels over Top-1 label performance, enabling the generation of more accurate candidate labels. For the second issue, Uni-LAP enhances the prediction capabilities of LLMs by incorporating syllogism-inspired reasoning, allowing the models to refine final predictions with greater logical coherence and alignment with the inherent complexity of legal articles.

We evaluate our proposed approach on two diverse, real-world datasets representing different regions and languages: the European Court of Human Rights (ECtHR) dataset and the Chinese AI and Law Challenge (CAIL2018) dataset. Empirical results demonstrate that Uni-LAP consistently outperforms all baseline methods, validating both the effectiveness and the generalizability of our approach.

The main contributions of this paper are as follows:
\begin{itemize}
    \item We explore the task of LAP from the perspective of collaboration between LLMs and SCMs. 
    \item We propose a universal approach to LAP (Uni-LAP) where the SCMs narrow the selection range using a novel Top-K loss function and the LLMs make the final prediction through syllogism-inspired reasoning.
    \item We conduct experiments on datasets from different regions, and the results on both datasets validate the effectiveness of the Uni-LAP.
    \item We make all the data and code publicly available to improve the reproducibility of our results\footnote{https://github.com/ZJULegalAI/Uni-LAP.git}.
\end{itemize}

This paper is organized as follows. Section 2 reviews related work. Section 3 specifies the problem, clarifies some terms of our framework, and introduces the Uni-LAP. Section 4 evaluates the model on two datasets by comparing it with the baseline models. Section 5 discusses the contributions, limitations, and future work.

\section{Related work}
\subsection{Legal Article Prediction}
Legal artificial intelligence (Legal AI) involves using AI technologies such as machine learning and NLP to automate legal tasks, assist decision-making, and improve the efficiency of legal processes. With the advancement of NLP techniques, more research in this field is focusing on assisting legal professionals with textual work, such as court view generation \cite{liu2024duapin} and legal question answering \cite{yao2025intelligent}. In this paper, we focus on LAP, a field that is becoming increasingly prominent in both legal AI and NLP \cite{DBLP:journals/access/CuiSW23}. The main goal of LAP is to predict legal articles from fact descriptions. Currently, this task is performed primarily by legal experts, which is both time-consuming and costly. With advances in deep learning, significant progress has been made in LAP \cite{DBLP:journals/ail/HouCZZ24}. Methods based on SCMs were the first to be developed \cite{DBLP:conf/sigir/LiuWZS0WK23, DBLP:conf/iciss2/StricksonI20}. These approaches, trained on specific datasets through supervised learning, aim to address specific tasks. More recently, the development of LLMs has introduced new approaches to LAP \cite{DBLP:conf/emnlp/DengMZD24, DBLP:conf/cikm/ShuZLDDZ24}. These models show strong capabilities in few-shot learning and contextual understanding, enabling them to tackle multiple tasks effectively. Despite these advancements, LAP remains a challenging task. LLMs often struggle with selecting the correct label due to the large number of abstract labels and prompt length constraints, while SCMs face challenges with confusing cases due to their limited scale. Recently, \citet{DBLP:conf/emnlp/WuZL0LZS0K23} explored legal judgment prediction by integrating LLMs and SCMs. However, their approach is limited to single-label classification and a single language, while also relying on an additional repository of similar cases, which constrains the model’s generalization ability. In contrast, our work emphasizes the close collaboration between SCMs and LLMs during SCM training and LLM reasoning. We enhance the model’s generalization capability by addressing multi-label text classification tasks, conducting experiments on both English and Chinese datasets, and eliminating the need for an additional case repository.

\subsection{Supervised Classification Models (SCMs)}
As NLP is widely applied in the legal domain, research on SCMs in this field has also increased. These models utilize annotated data to capture the complexity of the legal domain and can effectively combine domain knowledge with customized datasets. In recent years, many studies have focused on fine-tuning general pre-trained language models (e.g., BERT) for legal judgment prediction. Two notable models are LegalBERT \cite{DBLP:journals/corr/abs-2010-02559}, trained on legal documents from the EU, UK, and US, and CrimeBERT \cite{zhong2019open}, pre-trained on crime data. However, the performance of these models is constrained by the size of the datasets and the parameters of the models, especially when dealing with complex cases and capturing complex facts. 

\subsection{Large Language Models (LLMs)}
With the growing attention on LLMs, researchers have recognized their strong abilities in legal reasoning and text generation. Their influence has extended beyond specific tasks to broader legal applications, such as LAP \cite{DBLP:conf/cikm/ShuZLDDZ24}. Compared to SCMs, LLMs, equipped with hundreds of millions or even billions of parameters and trained on massive text corpora, demonstrate better capabilities in natural language understanding and handling of complex linguistic tasks. 

In the context of LAP, the performance of LLMs largely depends on the quality of their training and the input data. Techniques such as in-context learning (ICL) \cite{DBLP:conf/emnlp/WuZL0LZS0K23} and retrieval augmented generation (RAG) \cite{feng2024cadlra} have shown great potential to improve the capabilities of LLMs within the LJP field. ICL allows models to use contextual information in the input to perform tasks without requiring additional training. To improve ICL effectiveness, research has focused on refining instruction generation and prompt design to improve the model’s ability to handle and reason through complex tasks \cite{DBLP:conf/emnlp/Dong0DZMLXX0C0S24}. RAG integrates information retrieval with text generation, aiming to improve the performance and accuracy of LLMs in domain-specific tasks. Currently, this technique has been applied in legal scenarios \cite{DBLP:conf/iccbr/WiratungaAJMMNWLF24,feng2024cadlra}. However, LLMs still struggle with LAP task, primarily due to prompt limitations that constrain their ability to effectively handle abstract labels. A promising approach to address this challenge is to integrate LLMs with SCMs.

\subsection{Legal Syllogism}
Legal syllogism, a fundamental form of deductive reasoning, derives conclusions from the logical relationships between major premises, minor premises, and conclusions. The application of syllogistic reasoning to law was first articulated by Ulpian, where he introduced the ``rule-fact-conclusion'' model \cite{d201915}. Over centuries, this method has evolved and become a cornerstone of legal analysis, facilitating the interpretation of statutes and the derivation of judicial decisions. In contemporary legal practice, legal syllogism serves as a standardized tool for ensuring the logical consistency and coherence of legal judgments, while also facing challenges related to the complexity and abstraction of legal rules.

Previous work in knowledge representation and reasoning has extensively employed syllogistic methods to enhance model inference capabilities, including symbolic reasoning and Bayesian inference \cite{prakken2013formalising, DBLP:journals/ail/Constant24}. 
% For instance, Prakken and Sartor utilized syllogistic methods within the ASPIC+ framework to bolster the logical consistency and inferential accuracy of arguments concerning the validity of legal norms \cite{prakken2013formalising}. Similarly, Axel Constant applied Bayesian networks to simulate legal syllogisms, optimizing logical consistency and argument reliability in reasoning about the relationship between legal rules and facts by computing posterior probabilities and entropy \cite{constant2024bayesian}. 
Currently, with the increasing application of LLMs in the legal domain, more research is exploring how to integrate syllogistic reasoning into legal judgment analysis. For example, Deng et al. employed a LLM based on legal elements matching, leveraging syllogistic reasoning to precisely extract and analyze legal elements from case facts \cite{deng2023syllogistic}. In this paper, we enhance the prediction capabilities of the LLM by transferring legal articles and fact descriptions into a major premise and a minor premise, respectively, to determine whether an article matched the case based on informal legal syllogism.

\section{Methodology}

\begin{figure*}[t]
\centering
\includegraphics[width=\textwidth]{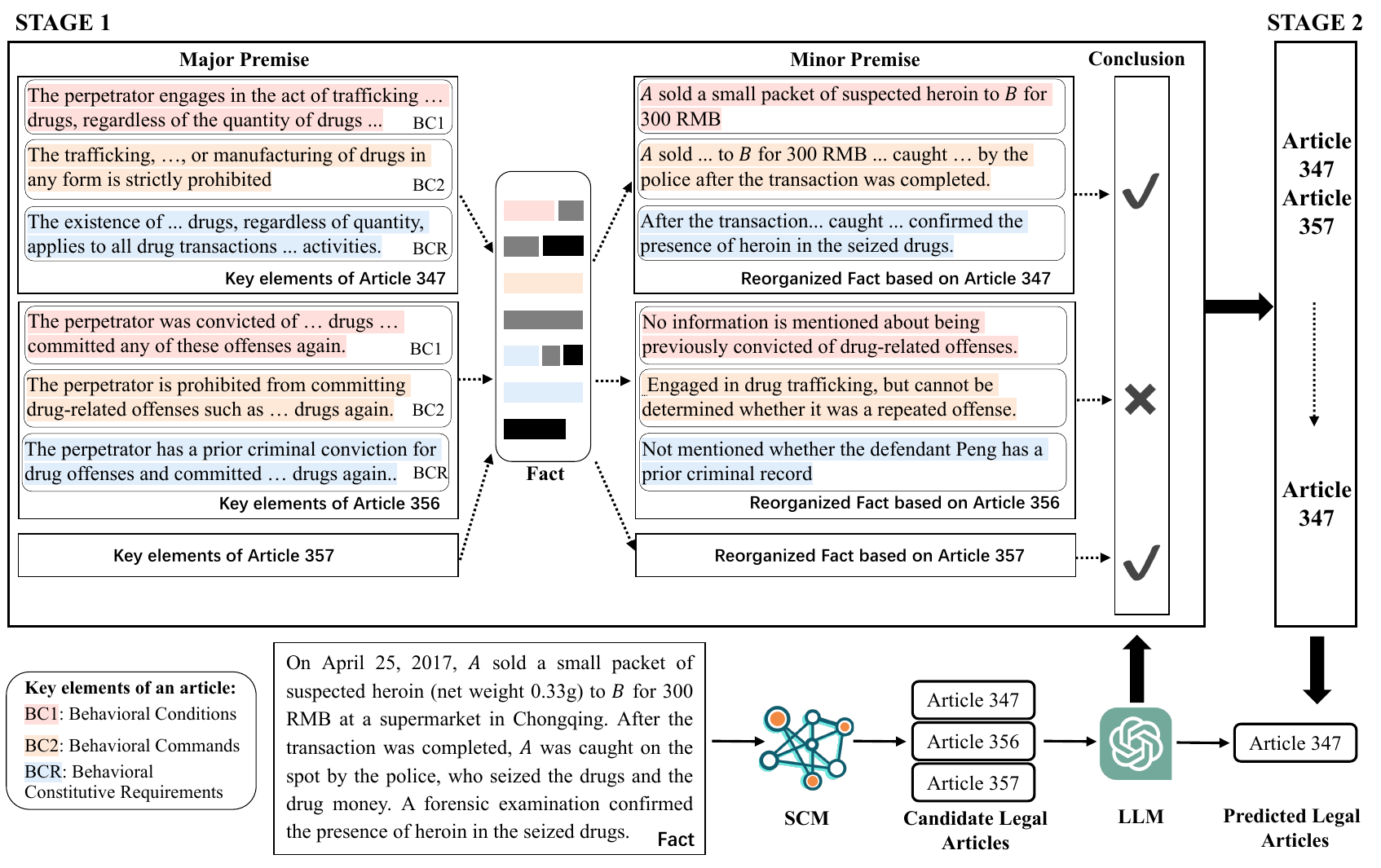}
\caption{The overall framework of Uni-LAP, where a check mark indicates `match', while a cross indicates `not match'. Solid lines represent the process of LAP, while the dotted lines represent the process inside the LLM. During the two calls of the LLM, ``Stage 1'' focuses on selecting matched articles while ``Stage 2'' focuses on selecting applicable articles from matched articles.}
\label{fig_overall}
\end{figure*}

In this section, we describe our universal approach to LAP (Uni-LAP). We first formulate LAP in this work as:

\begin{problem}[Legal Article Prediction]
    Given the fact description $f$, our task is to predict a set of candidate legal articles $A_{c}$ and select the final legal articles $A_{f}$ from them.
\end{problem}

We then clarify the definitions of some key terms.
\begin{itemize}
    \item \textbf{Fact description} refers to the detailed narration of the events and circumstances that are pertinent to the case, and is defined as a word sequence $f = \{w_{t}^{f}\}^{l_{f}}_{t=1}$, where $l_{f}$ is the total number of tokens. 
    \item \textbf{Candidate law articles} $A_{c}$ is a set of labels $\{c_{1},...,c_{k}\}$, where each $c_{i}$ represents a legal article predicted by the SCM, and $k=3$ in this paper.
    \item \textbf{Final law articles} $A_{f}$ is a set of labels $\{f_{1},...,f_{m}\}$, $m \leq k$, where each $f_{i}$ represents a legal article selected by the LLM.
\end{itemize}

Our Uni-LAP consists of two stages: SCM training and LLM reasoning. The overall framework is illustrated in Figure \ref{fig_overall}.

\subsection{Training of SCM}
The main task of the SCM is to select a set of candidate labels from hundreds of legal articles. This task concentrates on the efficiency of label generation and the quality of the selected candidate labels, rather than on the accuracy of the final prediction. In our work, the SCM identifies a candidate set containing the top three labels with the highest probabilities for further evaluation by the LLM. Therefore, the quality of the candidate articles is critical for the performance of the LLM. In order to predict higher-quality candidate sets of articles, the SCM training process is customized to facilitate better collaboration with the LLM in subsequent processes. Specifically, the process has two key steps: generation of candidate articles based on the probability distribution of legal articles and generation optimization using a TopK-Loss and a joint loss function.

\begin{figure}
\centering
\includegraphics[width=0.45\textwidth]{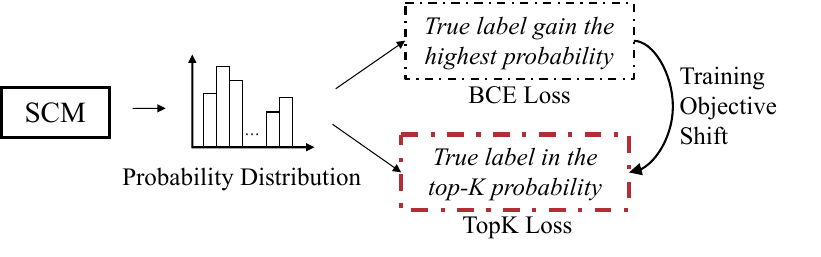}
\caption{The training objective of SCM.}
\label{fig_SCM}
\end{figure} 

\noindent \textbf{Candidate Articles Generation}
To generate a probabilistic distribution, the framework first transforms the fact description $f = \{w_{t}^{f}\}^{l_{f}}_{t=1}$ into an embedding sequence $H^{f} \in \mathbb{R}^{l_{f} \times d}$ through an encoder, expressed as:
\begin{equation}
    H^{f} = Encode(f),
\end{equation}
where $d$ represents the embedding dimension and $H^{f} = h_{1}^{f},...,h_{l_{f}}^{f}$ contains contextual information at all time steps $t$. Then we process $H_{f}$ through an feature extraction module to capture the overall semantic information of the fact description, producing a global feature representation $h^{f} \in \mathbb{R}^{d}$. Finally, we obtain the probabilistic distribution $P \in \mathbb{R}^{m}$ by processing the global feature representation $h^{f}$ through a fully connected network with sigmoid activation as follows:
\begin{equation}
    P = \text{Sigmoid}(W^{p} \cdot h^{f}+b^{p}),
\end{equation}
where $W^{p} \in \mathbb{R}^{m \times d}$ and $b^{p}\in \mathbb{R}^{m}$ are learnable parameters, and $m$ is the total number of legal articles.
% use a max-pooling operation to aggregate the embedding vectors at each time step into a global feature representation:
% \begin{equation}
%     h^f = \text{MaxPooling}(H^f)
% \end{equation}
% where $h^{f} \in \mathbb{R}^{d}$, thereby capturing the overall semantic information of the fact description. 
% Finally, we obtain the probabilistic distribution $P \in \mathbb{R}^{m}$ by processing the global feature representation $h^{f}$ through a fully connected network with softmax activation as follows:
% \begin{equation}
%     P = \text{Softmax}(W^{p} \cdot h^{f}+b^{p}),
% \end{equation}
% where $W^{p} \in \mathbb{R}^{m \times d}$ and $b^{p}\in \mathbb{R}^{m}$ are learnable parameters, and $m$ is the total number of legal articles.
% % and $P_{j}$ denotes the predicted probability of the $j$-th legal article. 
Based on the probabilistic distribution $P$, we selected the Top-K legal articles with the highest probabilities to form the initial candidate set of legal articles, denoted as $A_{c} = \{c_{1},c_{2},...,c_{k}\}$.

\noindent \textbf{Generation Optimization with TopK-Loss}
As Figure \ref{fig_SCM} shows, the goal of this step is to optimize the candidate set by ensuring that the true legal articles are at most included in the candidate set $A_{c}$ while minimizing interference from irrelevant legal articles. This step is specifically designed to address the challenge of the LLM being unable to handle all legal articles at once. To achieve this goal, we design a specialized loss function from two perspectives. 

The first perspective focuses on improving the precision of the predicted probability for each label. To achieve this, the classical Binary Cross-Entropy (BCE) loss function is employed. The BCE loss function optimizes the predicted probabilities $\hat{y}_{ij}$ for each label $j$ and each sample $i$, based on the corresponding ground truth labels $y_{ij}$. The BCE loss is shown as follows:

\begin{equation}
    \mathcal{L}_{BCE} = - \frac{1}{n} \sum_{i=1}^{n}\sum_{j=1}^{m} \left[ y_{ij} \cdot \log(\hat{y}_{ij}) + (1 - y_{ij}) \cdot \log(1 - \hat{y}_{ij}) \right],
\end{equation}
where $n$ is the total number of samples, $m$ is the total number of labels (e.g. legal articles), $y_{ij}$ is the ground truth label (either 0 or 1), and $\hat{y}_{ij}$ is the predicted probability for label $j$ of sample $i$.

The second perspective focuses on ensuring that the true legal articles are included in the candidate set, thereby improving the overall quality of the selected candidates while balancing efficiency and accuracy in candidate generation. To achieve this, we propose a TopK relaxation loss function (TopK-Loss) that evaluates the extent to which the true label set is covered by the candidate set. Specifically, the function evaluates whether the Top-K candidate set selected by the model covers the true label set, assigning a penalty when true labels are missing from the Top-K predictions, weighted by their predicted probabilities and importance.

The TopK-Loss is defined as follows:
%需要修改
\begin{equation}
    \mathcal{L}_{\text{TopK}} = \frac{1}{n} \sum_{i=1}^{n} \sum_{j=1}^{m} y_{ij} \cdot \hat{y}_{ij} \cdot (1 - M_{ij}),
\end{equation}
where $M_{ij}$ is an indicator function that determines whether class $j$ is among the Top-K predicted labels for sample $i$, defined as:
% where $n$ is the total number of samples, $\ell$ is the total number of label categories, $T_{ij}$ represents the ground truth label of sample $i$ for category $j$, and $\hat{Y}_{ij}$ denotes the softmax-normalized predicted probability of sample $i$ for category $j$. $M_{ij}$ is an indicator function that determines whether class $j$ is among the Top-K predicted labels for sample $i$, defined as:
\begin{equation}
    M_{ij} =
\begin{cases}
1, & \text{if } j \in \text{TopK}(\hat{y}_{i}) \\
0, & \text{otherwise}
\end{cases}
,
\end{equation}
where $\text{TopK}(\hat{y}_{i})$ represents the set of Top-K labels with the highest predicted probabilities from the predicted probability vector $\hat{y}_{i}$.

% $$
% \mathcal{L}_{\text{Top-K}} =
% \begin{cases} 
% 0, & \text{if } y^* \in \text{Top-K}(\hat{y}) \\
% 1, & \text{if } y^* \notin \text{Top-K}(\hat{y})
% \end{cases}
% ,
% $$
% where $y^{*}$ denotes the true legal article(s), $\hat{y}$ denotes the predicted probability distribution for all labels, the set $Top-K(\hat{y})$ includes the K labels with the highest predicted probabilities.

Based on the BCE loss function and the TopK-Loss, the overall loss function is defined as follows:
\begin{equation}
    \mathcal{L}_{\text{overall}} = \lambda_1 \mathcal{L}_{\text{BCE}} + \lambda_2 \mathcal{L}_{\text{TopK}},
\end{equation}
where $\lambda_1$ and $\lambda_2$ are the weight parameters.

In this way, the overall loss function simultaneously optimizes both prediction accuracy and the coverage of the candidate set.

\subsection{LLM Reasoning}
LLMs are trained on large-scale corpora and equipped with billions of parameters. Therefore, they can handle tasks that require deep understanding and thoughtful decision-making. Due to their strong reasoning abilities, the main task of the LLM in our framework is to conduct reasoning based on the candidate set of legal articles predicted by the SCM and select the final applicable articles.\footnote{Note that when we refer to `legal reasoning', 
our intention is not to suggest that LLMs themselves are capable of 
reasoning, but rather to indicate that their outputs reflect steps from 
the judicial reasoning process.} To achieve better results, our overall prompt is divided into two main stages, with the LLM being called twice during the process. The first stage aims to select legal articles that are considered to be matched. However, these articles are not necessarily the applicable legal articles for a case. For example, the articles on illegal intrusion and intentional property damage may both match a case, but as the focus is on significant property damage, the latter is the applicable one. Therefore, the goal of the second stage is to identify the most relevant legal article among all those matched articles.

\begin{figure}
\centering
\includegraphics[width=0.25\textwidth]{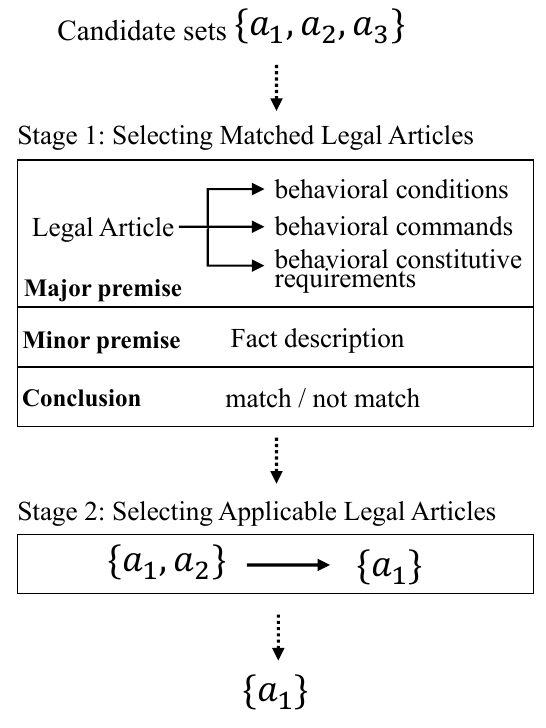}
\caption{The reasoning process of the LLM, where $\{a_{1},a_{2},a_{3}\}$ represents the candidate set predicted by the SCM, $\{a_{1},a_{2}\}$ denotes the set of matched articles, and $\{a_{1}\}$ is the final predicted article for the case. The solid lines represent the overall prompt reasoning process, while the dotted lines represent the internal process of each stage.}
\label{fig_LLM}
\end{figure} 

\subsubsection{Stage 1}
% \noindent \textbf{Stage 1}
The objective of stage 1 is to improve the matching process between legal articles and fact descriptions. A useful approach is to analyze the logical structure of legal articles systematically and break them down into specific key elements. The breakdown provides a structured way to clarify the applicable conditions and the core content of legal articles, thereby improving the accuracy and interpretability of the matching process. The legal syllogism is a liberal form of categorical syllogism whereby an implication relation between a minor premise and a major premise is used to infer a conclusion \cite{DBLP:journals/ail/Constant24}. Inspired by the syllogism, in our reasoning process, the major premise serves as the theoretical basis, clarifying the applicable conditions and behavioral requirements of a legal article. It provides legal context for a specific case. For LAP, legal consequences are not included in the major premise, as conclusions have not yet been reached when determining whether a legal article matches a fact. The minor premise provides a factual basis for the reasoning process. The conclusion involves inferring the matched legal articles based on the major and minor premises. Note that we only draw on the intuition of legal syllogism and apply it informally, without strictly adhering to its formal structure.

Based on the structure of syllogism, the overall reasoning process of stage 1 is divided into the following three key steps and is carried out using prompts for the LLM.

%In practice, syllogistic reasoning is employed by judges to maintaining consistency and fairness in their decision-making process \cite{hart2012concept}.

\vspace{0.5em}
\noindent\textbf{(1) Major Premise} \ \
In this step, we aim to provide legal context for specific cases. More specifically, we aim to extract the key elements of legal articles. The prompt is designed according to \cite{Shu2022}, where we believe that a complete legal article should logically consist of three key elements except for legal consequences, i.e., behavioral conditions, behavioral commands, and behavioral constitutive requirements. Behavioral conditions are the prerequisites for applying the legal article and describe the qualifications that actors must meet and the circumstances under which their actions occur. For example, certain legal articles may require the actor to possess a specific identity or for the actions to take place in a particular situation. Behavioral commands specify the requirements for the actor’s actions and include three forms: mandatory actions, prohibited actions, and permissible actions. Behavioral constitutive requirements refer to the `factual patterns’ prescribed by the legal article, which are specific behavior patterns or events. 
% Behavior conditions are the prerequisites for applying the legal article; behavior commands specify the requirements for actor’ actions; behavior constitutive requirements refer to the ‘factual patterns’ prescribed by the legal article; and legal consequences describe the outcomes when the legal article is applied. 
Based on above, Prompt for step $1$ is designed as follows:\ \
``\textit{Please extract the following key elements from the legal article: behavioral conditions, behavioral commands, and behavioral constitutive requirements. Behavioral conditions describe the prerequisites for applying the legal article, including the qualifications of the actor and the specific context in which the behavior occurs; behavioral commands clarify the requirements that the actor must follow and can be categorized as `mandatory actions’, `prohibited actions’, and `permissible actions’; behavioral constitutive requirements specify the particular behavior patterns or events defined by the legal article.''}

\vspace{0.5em}
\noindent\textbf{(2) Minor premise} \ \
The goal of this step is to guide the model to reorganize the fact descriptions based on the key elements of the major premise. To achieve this goal, the Prompt for step $2$ is designed as follows:\ \
``\textit{According to the fact description, please identify the text segments related to the three key elements, i.e., behavioral conditions, behavioral commands, and behavioral constitutive requirements.}''

% Our goal in this step is to guide the model to gradually filter and compare the fact descriptions with the key elements extracted from the legal article. To achieve this goal, Prompt $2$ is designed as follows:\ \
% ``\textit{Based on the extracted key elements from the major premise, please identify the parts of the provided fact descriptions that align with the behavior conditions, behavior commands, behavior constitutive requirements, and legal consequences.}''

\vspace{0.5em}
\noindent\textbf{(3) Conclusion} \ \
The LLM gradually determines whether the fact descriptions align with the elements of a legal article based on the aforementioned major premise and minor premise, and ultimately decides whether the article is matched. Specifically, if the fact descriptions meet all the conditions of the legal article, the conclusion is that the article is matched; otherwise, the article is not matched. The prompt for step $3$ is presented as follows:\ \
``\textit{First, compare the text segments related to the three key elements with the corresponding elements in the legal article. Specifically, as follows: First, compare the behavioral conditions in the fact descriptions with those defined in the legal article to determine whether the facts fall within the specified context of the article. Next, check whether the fact descriptions involve behaviors prohibited, required, or permitted by the legal article. Third, verify whether the behaviors in the fact descriptions constitute the behavioral constitutive requirements in the legal article. It is important to note that when examining the fact descriptions, you should check whether the fact descriptions match the content of the legal article in terms of semantics rather than relying solely on literal similarity. Second, evaluate whether the legal article matches the case. The conclusion must be either `match’ or `not match’. Please assess whether an article matches the case according to the following criteria: If the fact descriptions meet all the conditions of the legal article (behavioral conditions, behavioral commands, and behavioral constitutive requirements), the article is matched. If any key element in the fact description does not align with the legal article, the article is not matched.}''

The prompts for the above three steps are combined into a single prompt and executed in one call to the LLM.

\subsubsection{Stage 2}
% \noindent \textbf{Stage2}
From stage 1, we can obtain legal articles that are considered to be matched. To predict the final applicable legal articles for a case, we design the prompt as follows:``\textit{Based on the matched legal articles and the specific legal analysis generated in Stage 1, the objective is to identify the most relevant legal articles. When two articles share the same scope of application, priority should be given to the one that appears earlier in the candidate set predicted by the SCM. If all identified articles are deemed irrelevant, the selection should then consider the entire set of criminal legal articles.}'' Since the LLM relies on zero-shot learning in our framework, there are situations when it may fail to identify the most suitable legal articles accurately. Therefore, utilizing the candidate set provided by the SCM can effectively improve its accuracy. Moreover, since the SCM predictions may not always be correct, the LLM is allowed to consider all legal articles when necessary.

\section{Experiments}

\begin{table}[ht]
\small
\centering
\resizebox{\columnwidth}{!}{
\begin{tabular}{lcc}
\hline
\textbf{Type} & \textbf{ECtHR} & \textbf{CAIL2018} \\ \hline
Language & English & Chinese \\
Domain & ECHR & Chinese Criminal Law \\
\# Cases& 11k & 186K \\
\# Classes & 10 & 73 \\
Avg. \# String Length in Fact & 10402.3  & 442.6 \\
Avg. \# Articles per case & 1.4  & 1.3 \\
\hline
\end{tabular}
}
\caption{Statistics of the datasets.}
\label{tab:datasets}
\end{table}

\begin{table*}[t]
\centering
\small % 缩小字体大小
\resizebox{0.8\textwidth}{!}{ % 调整表格宽度
\begin{tabular}{lcccc|cccc}
\hline
\multirow{2}{*}{\centering \textbf{Method}} & \multicolumn{4}{c|}{\textbf{ECtHR}} & \multicolumn{4}{c}{\textbf{CAIL2018}} \\
\cline{2-9}
 & \textbf{Acc} & \textbf {Ma-F} & \textbf{Ma-P} & \textbf{Ma-R} & \textbf{Acc} & \textbf{Ma-F} & \textbf{Ma-P} & \textbf{Ma-R} \\
\hline
CNN \cite{DBLP:journals/neco/LeCunBDHHHJ89} & 33.16 & 39.42 & 41.70 & 37.76 & 68.49 & 77.57 & 79.40 & 76.95 \\
RNN \cite{DBLP:journals/cogsci/Elman90} & 36.61 & 52.35 & 51.74 & 53.97 & 71.28 & 79.22 & 76.77 & 83.09 \\
LegalBERT \cite{DBLP:journals/corr/abs-2010-02559} & 58.01 & \underline{77.24} & 77.45 & 77.90 & \textemdash & \textemdash & \textemdash & \textemdash \\
CrimeBERT \cite{zhong2019open} & \textemdash & \textemdash & \textemdash & \textemdash & 74.99 & 82.41 & 80.30 & \underline{85.66} \\
\hline
gpt-4o & 21.57 & 65.68 & 54.92 & \textbf{85.51} & 57.98 & 60.20 & 66.02 & 65.46 \\
gpt-3.5-turbo & 8.82 & 50.25 & 44.76 & 74.66 & 0.40 & 5.66 & 5.83 & 12.04 \\
\hline
Uni-LAP (CNN+gpt-4o) & 72.72 & 73.56 & \underline{82.64} & 68.81 & 85.68 & 84.88 & 86.95 & 84.31 \\
Uni-LAP (RNN+gpt-4o) & \underline{74.95} & 76.29 & 78.59 & 74.69 & \underline{86.14} & \underline{85.18} & \underline{87.34} & 84.03 \\
Uni-LAP (LegalBERT+gpt-4o) & \textbf{83.16} & \textbf{83.20} & \textbf{84.27} & \underline{82.30} & \textemdash & \textemdash & \textemdash & \textemdash \\
Uni-LAP (CrimeBERT+gpt-4o) & \textemdash & \textemdash & \textemdash & \textemdash & \textbf{87.56} & \textbf{87.26} & \textbf{89.52} & \textbf{86.33} \\
Uni-LAP (LegalBERT+gpt-3.5-turbo) & 67.29 & 68.65 & 70.15 & 71.75 & \textemdash & \textemdash & \textemdash & \textemdash \\
Uni-LAP (CrimeBERT+gpt-3.5-turbo) & \textemdash & \textemdash & \textemdash & \textemdash & 82.50 & 78.05 & 82.60 & 77.77 \\
\hline
\end{tabular}}
\caption{Results of different methods on the ECtHR and CAIL2018 datasets. The best is \textbf{bolded}, and the second best is \underline{underlined}.}
\label{tab:results}
\end{table*}

\subsection{Datasets}
We experiment with two multi-class LAP datasets from different regions in two languages, as shown in Table \ref{tab:datasets}.

\noindent\textbf{ECtHR} \ \
The European Court of Human Rights (ECtHR) hears allegations that a state has breached the human rights provisions of the European Convention of Human Rights (ECHR) \cite{DBLP:conf/acl/ChalkidisAA19}. In our experiments, we use the dataset ECtHR Tasks B \cite{DBLP:conf/naacl/ChalkidisFTAAM21}, where the input is a list of facts of a case and the output is a set of allegedly violated articles\footnote{For simplicity, in the following paper, we refer to ECtHR Task B as ECtHR.}. 
% This dataset consists of real-world cases, each of which includes a factual description and a binary label indicating whether a specific ECHR article was violated.
% The cases are chronologically split into training (80k, 2013-2017), validation (12k, 2017-2018), and test sets (12k, 2018).
We performed data cleaning on the dataset by removing sequence numbers and spaces at the beginning of the samples and excluding samples with empty labels. After cleaning, the dataset retained 8,866 samples for the training set, 973 samples for the validation set, and 986 samples for the test set.

\vspace{0.5em}
\noindent\textbf{CAIL} \ \
The Chinese AI and Law Challenge dataset (CAIL) comprises a series of datasets from the People’s Republic of China (PRC). We use CAIL-small of the CAIL2018 dataset \cite{DBLP:journals/corr/abs-1807-02478}, where each case includes a fact description and its corresponding legal judgment. The result of the judgment of each case is refined into relevant articles, charges, and prison terms, and we focus on the legal articles. 

We performed data cleaning to address label imbalance by retaining only legal articles with more than 1,000 appearances and obtained 73 high-frequency articles. Additionally, we removed duplicate legal articles in each case to ensure unique labels. After cleaning, the dataset retained 140,789 samples for the training set, 15,475 samples for the validation set, and 29,388 samples for the test set. 
% To handle text inputs, we set the maximum number of tokens to 510, which is sufficient for the average token length required by the dataset. 
Furthermore, to prevent data leakage, we filtered out cases where the charges or legal articles have been leaked, as the LLM might use known information to infer applicable legal articles, which could potentially impact its predictive performance. 

\subsection{Baselines}
For SCM baselines, we implement the following for comparison: 

\textbf{CNN} \cite{DBLP:journals/neco/LeCunBDHHHJ89} extracts text features through convolutional operations with different kernels for text classification. \textbf{RNN} \cite{DBLP:journals/cogsci/Elman90} are designed to handle sequence data by maintaining a hidden state that captures information about previous elements in the sequence. \textbf{LegalBERT} \cite{DBLP:journals/corr/abs-2010-02559} is a BERT model pre-trained on legal corpora from the EU, UK, and US. \textbf{CrimeBERT} \cite{zhong2019open} is a model initialized from BERT and further pre-trained on crime data, achieving better performance in legal judgment prediction tasks. Particularly, LegalBERT has been pre-trained on ECHR cases, while CrimeBERT has been pre-trained on Chinese criminal cases. Therefore, in the experiments, we implement LegalBERT for ECtHR and CrimeBERT for CAIL2018 to achieve better performance
\footnote{To handle long texts in the ECtHR dataset, we incorporated HierarchicalBert into the LegalBERT model (as referenced in LexGLUE \cite{DBLP:conf/acl/ChalkidisJHBAKA22}).}.
% \textbf{Hierarchical BERT} \cite{DBLP:conf/acl/ChalkidisJHBAKA22} is a BERT-based model designed to encode facts and legal articles using a hierarchical architecture to handle long legal documents effectively.

For LLM baseline, we use \textbf{gpt-4o} and \textbf{gpt-3.5-turbo}, language models released by OpenAI \footnote{https://platform.openai.com/docs/models}, designed for general applications.

% For our method, we take.. 
% We do ablation experiments as follows: \textbf{Uni-LAP w/o k} refers to the removal of the $TopK$ loss function from the SCM; \textbf{Uni-LAP w/o t1} refers to the removal of stage 1, i.e., syllogism-inspired reasoning, from the reasoning process of the LLM, so that the LLM makes predictions based solely on its ordinary understanding; \textbf{Uni-LAP w/o t2} refers to the removal of stage 2 from the reasoning process of the LLM, so that the LLM makes predictions based solely on the matched candidate labels; \textbf{Uni-LAP w/o l} means that the LAP task is performed solely by the SCM.

\subsection{Experiment Setup}
In this section, we describe the implementation details of Uni-LAP in our experiment. The LAP task in this work is a multi-label classification task, which means each case can have more than one applicable law article.

For the SCMs, we use CNN, RNN, and LegalBERT for ECtHR, and CNN, RNN, and CrimeBERT for CAIL2018. In our experiment, we set the SCM threshold to 0.3. That is if the probability of a certain label exceeds 0.3, it is selected out, indicating a positive outcome. 

To facilitate the validation of experiments, we set the candidate set length \textbf{K=3}, and this setup has minimal impact on the performance of our experiment due to the following reasons. For the ECtHR, the average number of articles per case is 1.4, with a median of 1.0 and a standard deviation of 0.6. Similarly, for the CAIL2018, the average number of legal articles per case is 1.3, with a median of 1.0 and a standard deviation of 0.8. These statistics indicate that the majority of samples in both datasets correspond to only one or a small number of legal articles. Furthermore, cases with more than three applicable legal articles account for only about $2\%$ of the dataset, with a maximum of $3\%$.

% The large model is called twice in total. The first call employs syllogism-inspired reasoning to analyze whether the three candidate labels meet the basic matching criteria. The second call selects the final legal article that best matches the case from the matched labels.

For the evaluation metrics, we employ accuracy (Acc), macro F1 (Ma-F), macro Precision (Ma-P), and macro Recall (Ma-R) to analyze the performance of our methods against the baselines. 

Additionally, we introduce a new metric, $TopK$ accuracy (TopK-ACC), to evaluate the performance of the supervised classification model in selecting candidate labels, and is defined as 

\begin{equation}
    \text{TopK-ACC} = \frac{\sum_{i=1}^{B} \sum_{j \in \mathcal{Y}_i} \mathbb{I}(j \in \mathcal{T}_i)}{\sum_{i=1}^{B} |\mathcal{Y}_i|},
\end{equation}
where $B$ is the batch size, $\mathcal{Y}_{i}= \{ j \mid Y_{i,j} = 1 \}$ is the set of true labels for the $i$-th sample, $\mathcal{T}_i$ is a Top-K predicted label set for the $i$-th sample, and $\lvert \mathcal{Y}_{i} \rvert$ is the number of true labels for the i-th sample.

\subsection{Experiment Results}
\subsubsection{Comparison Against Baselines}
To evaluate the effectiveness of Uni-LAP, we collaborate gpt-4o with each baseline model that incorporates the TopK-Loss. Additionally, we include a collaboration of gpt-3.5-turbo with LegalBERT and CrimeBERT (also using the TopK-Loss) on the ECtHR and CAIL2018 datasets, respectively, to compare the LLMs. The results of our main experiment are listed in Table \ref{tab:results}. The performance results of the SCMs in selecting candidate labels are shown in Table \ref{tab:topk_results}.

\begin{table}[t]
\centering
\small % Reduce font size
\resizebox{0.32\textwidth}{!}{ 
\begin{tabular}{lcc}
\hline
\multirow{2}{*}{\centering \textbf{Method}} & \multicolumn{1}{c}{\textbf{ECtHR}} & \multicolumn{1}{c}{\textbf{CAIL2018}} \\
\cline{2-3}
 & \textbf{TopK-ACC} & \textbf{TopK-ACC} \\
\hline
CNN & 75.74 & 81.80 \\
+TopK-Loss & \textbf{79.58\textuparrow} & \textbf{92.51\textuparrow} \\
\hline
RNN & 78.54 & 91.31 \\
+TopK-Loss & \textbf{80.98\textuparrow} & \textbf{92.41\textuparrow} \\
\hline
LegalBERT & 88.78 & \textemdash \\
+TopK-Loss & \textbf{92.82\textuparrow} & \textemdash \\
\hline
CrimeBERT & \textemdash & 93.42 \\
+TopK-Loss & \textemdash & \textbf{94.12\textuparrow} \\
\hline
\end{tabular}}
\caption{TopK-ACC of different SCMs.}
\label{tab:topk_results}
\end{table}

From Table \ref{tab:results}, we have the following observations. 
(1) The SCMs do not perform well in the LAP task, especially on the ECtHR. This is mainly because cases in ECtHR are much longer than those in CAIL and contain many long sentences. As a result, SCMs are constrained by their scale and struggle to effectively handle long texts and the confusing parts of facts, including uncommon legal terms in generic corpora (e.g., `novation', `pillory effect'), terms with meanings that differ from everyday usage (e.g., `removal of children’ in a legal context refers to government intervention and the deprivation of parental custody), and long sentences with unusual word order (e.g., ``since, in the present case, it was the child’s biological mother who did not know how to respond or touch the child, the contact experience could be traumatic for E''). 
(2) The LLMs, i.e., gpt-4o and gpt-3.5-turbo, perform poorly in the LAP task on both datasets. This is because, without candidate labels, the LLM must select the correct legal articles from the entire label set, which poses a significant challenge in handling a large number of abstract labels due to its prompt-based usage, leading to an over-reliance on superficial features, insufficient semantic understanding, and limited generalization capabilities. 
% (3) Among LLMs, gpt-4o performs better on the LAP task.
(3) We found that gpt-4o achieves the highest macro Recall score on ECtHR but has very low accuracy. The same issue occurs with gpt-3.5-turbo. This is due to the imbalanced sample distribution in the ECtHR dataset, where some classes have significantly more samples than others. As a result, the model tends to predict the majority class, leading to a high recall rate (i.e., capturing positive samples from the majority class) but low accuracy (i.e., poor performance on the minority class).
(4) The LAP in this work is a multi-label text classification task, so we focus on analyzing the accuracy and macro F1 scores for both models. For Uni-LAP based on LegalBERT, accuracy increased by $25.15\%$ and macro F1 score increased by $5.96\%$ on ECtHR. For Uni-LAP based on CrimeBERT, accuracy increased by $12.57\%$ and Macro F1 score increased by $4.85\%$ on CAIL2018. These results indicate that our framework can accurately predict the legal articles corresponding to cases and demonstrates strong predictive capability across all classes.
(5) By cooperating the SCMs with the LLMs, the performance of all baseline models significantly improved, achieving state-of-the-art results on both datasets, demonstrating the effectiveness of our framework. Among them, Uni-LAP (CrimeBERT+gpt-4o) achieves the best performance.

From Table \ref{tab:topk_results}, we can observe that:
(1) applying the TopK-loss significantly enhances the TopK-ACC performance across all baseline models, demonstrating the effectiveness of our proposed TopK-loss in selecting better candidate articles
(2) For simple SCMs such as CNN and RNN, the improvement is more significant.

\subsubsection{Ablation Study}

% \textbf{Uni-LAP w/o t1} refers to the removal of stage 1 from the reasoning process of the LLM, so that the LLM makes predictions based solely on its ordinary understanding; \textbf{Uni-LAP w/o t2} refers to the removal of stage 2 from the reasoning process of the LLM, so that the LLM makes predictions based solely on the matched candidate labels; \textbf{w/o k} refers to the removal of the Top-K-Loss from the SCM when predicting candidate labels. 

\begin{table}[t]
\centering
\small % 缩小字体大小
\resizebox{0.45\textwidth}{!}{ % 调整表格宽度
\begin{tabular}{lcc|cc}
\hline
\multirow{2}{*}{\centering \textbf{Method}} & \multicolumn{2}{c|}{\textbf{ECtHR}} & \multicolumn{2}{c}{\textbf{CAIL2018}} \\
\cline{2-5}
 & \textbf{Acc} & \textbf {Ma-F} & \textbf{Acc} & \textbf{Ma-F}\\
\hline
Uni-LAP w/o k     & 82.06 & 82.71          & 85.40 & 85.83 \\
Uni-LAP w/o $t_{1}$ & 82.29 & \textbf{83.85} & 86.21 & 85.58 \\
Uni-LAP w/o $t_{2}$ & 28.68 & 69.36 & 59.45 & 80.72 \\
\hline
Uni-LAP &\textbf{83.16} & 83.20 & \textbf{87.56} & \textbf{87.26} \\
\hline
\end{tabular}}
\caption{Results of ablation experiments, where Uni-LAP w/o $k$ refers to the removal of the TopK-Loss from the SCMs, Uni-LAP w/o $t_{1}$ refers to the removal of stage 1 from the reasoning process of the LLM, and Uni-LAP w/o $t_{2}$ refers to the removal of stage 2 from the reasoning process of the LLM.}
\label{tab:ablation_llm}
\end{table}

From Table \ref{tab:ablation_llm}, we can draw the following conclusions:
(1) The performance difference between Uni-LAP w/o $k$ and Uni-LAP on both datasets underscores the effectiveness of the TopK-Loss in LAP.
(2) For CAIL2018, the performance gap between Uni-LAP w/o $t_{1}$ and Uni-LAP demonstrates the importance of stage 1 in the LLM reasoning process for this dataset. For ECtHR, the slightly higher macro F1 score of Uni-LAP w/o $t_{1}$ is primarily attributed to class imbalance in the dataset. Since Uni-LAP w/o $t_{1}$ exhibits weaker reasoning capabilities, it may produce more random predictions for some minority classes, which obscures its underlying limitations.
(3) For both datasets, stage 2 in the LLM reasoning is relatively important due to the significant performance gap between Uni-LAP w/o $t_{2}$ and Uni-LAP on both datasets.

\subsubsection{Error Analysis}

% We analyzed the error cases of our method, and found that the errors are mainly due to the following three types of reasons: 
% excluding those caused by actual label inaccuracies, label leakage, SCM prediction errors, or the involvement of sensitive words. 
We analyzed the error cases of the reasoning of LLMs, which can be classified into two categories:

$\bullet$ Stage 1 error: These errors occur in the first stage of the LLM reasoning process. In CAIL2018, they account for 56.2$\%$ of LLM-induced errors and 22.8$\%$ of total errors. In ECtHR, they account for 18.8$\%$ of LLM-induced errors and 10.3$\%$ of total errors. The errors occur due to two main reasons. First is that the model cannot clearly distinguish the applicable scope of keywords in some similar legal articles, e.g., `toxic and harmful food' versus `food not meeting safety standards'. Second is that there exist additional regulations for determining relationships between certain related legal articles, and the LLM cannot correctly determine whether the legal articles match a case in such situations. 

$\bullet$ Stage 2 error: These errors occur in the second stage of the LLM reasoning process. In CAIL2018, they account for 43.8$\%$ of LLM-induced errors and 22.8$\%$ of total errors. In ECtHR, they account for 81.2$\%$ of LLM-induced errors and 17.7$\%$ of total errors. The errors occur because some facts are associated with multiple legal articles that have only minor differences in relevance. Specifically, these errors can arise for two reasons. One is that the LLM fails to correctly assess the relevance of certain articles, the other is that articles are open-textured, meaning articles beyond the ground-truth labels may also be applicable to the case.

\begin{figure}
\centering
\includegraphics[width=0.45\textwidth]{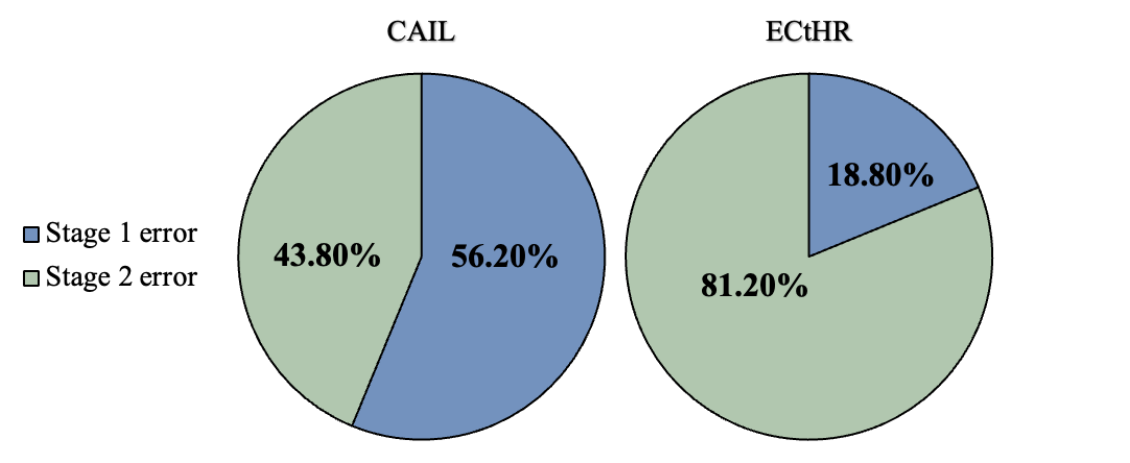}
\caption{Proportion of LLM-induced errors.}
\label{fig_error}
\end{figure}

As shown in Figure \ref{fig_error}, we can observe that for the CAIL dataset, the error proportions of the two stages of LLM reasoning are relatively balanced. However, for the ECtHR dataset, the proportion of errors in Stage 2 is significantly higher than in Stage 1. This is primarily because the fact descriptions in ECtHR are often long texts, containing rich but complex information. Additionally, Article 6 (Right to a Fair Trial), which is highly relevant to this dataset, is broadly associated with numerous cases. As a result, in Stage 1, the model can easily identify matched articles based on whether the fact descriptions align with the key elements of a legal article, as the rich information in long texts increases the likelihood of matches. However, in Stage 2, the model faces greater difficulty when determining whether these matched articles are truly applicable, primarily due to the contextual complexity and the ambiguous associations inherent in long texts.

In the future, to address Stage 1 errors, the text can be segmented with contextual connections established between segments. To address Stage 2 errors, additional annotations and explanations can be added to the legal dataset.

%定性 统一分析
%定量 分别统计一下
%哪些点可以提升

\subsection{Case Study}

\begin{figure}
\centering
\includegraphics[width=0.45\textwidth]{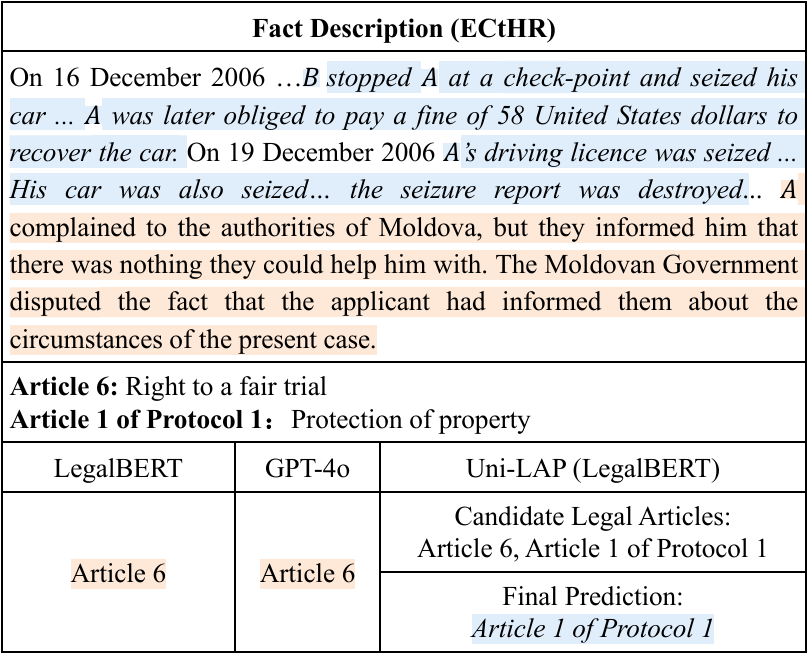}
\caption{The article prediction of a given case on ECtHR, predicted by LegalBERT, gpt-4o, and Uni-LAP respectively, where the \textit{blue} parts contain information aiding both the SCM and the LLM in predicting articles, while the orange parts provide additional information for Uni-LAP.}
\label{fig_CaseECtHR}
\end{figure}

In this section, we will use a real-world case to demonstrate the effectiveness of our framework in LAP, as shown in Figure \ref{fig_CaseECtHR}. It presents the predicted legal articles for a case by different models. In this case, both articles are relevant, and the models need to select the most appropriate one. Both SCM and LLM focused on the orange content (emphasizing that the applicant did not receive legal remedies) and predicted Article 6 (the right to a fair trial) as the correct one. In contrast, Uni-LAP additionally considered the blue content (highlighting the unlawful seizure of $A$’s car), identified it as the main issue of the case, accurately assessed the applicability of the open-textured legal term `deprivation', and correctly predicted that Article 1 of Protocol 1 was more suitable.

Both SCM and LLM incorrectly determined that the case mainly concerns Article 6 (the right to a fair trial) due to the context, `the applicant did not receive legal remedies', although fair trial rights might be a secondary consideration. Alternatively, Uni-LAP accurately identified that the main issue is the `unlawful seizure of $A$'s car' and predicted that Article 1 of Protocol 1 is more suitable. 

The following is a detailed legal analysis generated by the LLM, based on syllogistic reasoning, illustrating how it follows a logical pattern to derive the final legal article. This analysis is well-structured and enhances the explainability of the LLM: \textit{For the ECtHR $Case\_2026$, we analyzed the applicability of Article 6: Right to a Fair Trial and Article 1 of Protocol 1: Protection of Property. Regarding Article 6, the case facts indicate that A sought judicial relief after his car and driving license were seized, but the Moldovan authorities failed to provide any effective remedy, disputing his complaints. This aligns with the behavioral conditions, obligations, and criteria outlined in Article 6, as A’s inability to access a fair trial and judicial review constitutes a violation of the right to a fair trial. Similarly, for Article 1 of Protocol 1, the facts reveal that A’s car and driving license, his legitimate properties, were forcibly seized without appropriate legal procedures or justification, and the seizure report was destroyed. These actions violate the protections of property ownership and constitute unjustified deprivation of property, satisfying the behavioral conditions, commands, and criteria of Article 1. While both articles apply to the case, Article 1 of Protocol 1 is more relevant due to its narrower focus on property protection. Therefore, the most applicable legal article is Article 1 of Protocol 1: Protection of Property.}

% \begin{figure*}[t]
% \centering
% \includegraphics[width=1.0\textwidth]{Uni-LAP/Figure/Case.jpg}
% \caption{The article prediction of a given case. The \underline{blue parts} contain information that aids both the SCM and the LLM in predicting articles, while the \dotuline{orange parts} provide additional information for Uni-LAP in predicting articles}
% \label{fig_Case}
% \end{figure*}
% Figure \ref{fig_CaseCAIL} presents the final predicted articles by CrimeBERT, gpt-4o, and Uni-LAP for a case in the CAIL2018 dataset. Both CrimeBERT and gpt-4o were misled by the phrase "pickpocketed a mobile phone" and incorrectly classified the case as theft instead of robbery. By collaborating the knowledge of abstract labels from the SCM with the knowledge from the LLM in the syllogism-inspired reasoning process, which thoroughly analyzes specific legal articles with the given facts, Uni-LAP identifies the presence of violent resistance in the facts. As a result, the charge is updated from theft to robbery, making Article 263 more appropriate for this case. 

\section{Conclusion and Future Work}
In this paper, we proposed Uni-LAP, a universal framework for LAP that harnesses the strengths of both SCM and LLM. We introduce a TopK-Loss to enhance SCM’s performance in predicting candidate articles and apply syllogism-inspired reasoning to strengthen LLM’s legal reasoning capabilities. Experimental results validate the effectiveness of Uni-LAP across multi-class legal datasets from different regions and languages.

In the future, it will be valuable to evaluate our model on datasets specifically designed for continental civil-law and common-law systems to enhance its effectiveness. Additionally, focusing on improving the transparency of the reasoning process in the LLM and improving feature interpretability in the SCM will help develop a more explainable LAP framework.

\section{Acknowledgements}
This work was supported by ``Pioneer'' and ``Leading Goose'' R\&D Program of Zhejiang (2025C02037), Ant Group,Chongqing Ant Consumer Finance Co, Zhejiang University Qiushi Eagle Plan.

% \input{Uni-LAP/1_Introduction}
% \input{Uni-LAP/2_Related_Work}
% \input{Uni-LAP/3_Methodology}
% \input{Uni-LAP/4_Experiments}
% \input{Uni-LAP/5_Conclusion}
% \input{Uni-LAP/6_Acknowledgements}

\bibliographystyle{ACM-Reference-Format}
\bibliography{sample-base}


\end{document}